\documentclass[10pt,twocolumn,letterpaper]{article}

\usepackage{wacv}
\usepackage{times}
\usepackage{epsfig}
\usepackage{graphicx}
\usepackage{amsmath}
\usepackage{amssymb}
\usepackage{booktabs}
\usepackage{makecell}
\usepackage[accsupp]{axessibility}
\def\wacvPaperID{1975} % Enter the WACV Paper ID here

\wacvalgorithmstrack   % Uncomment this line if you are submitting to the Algorithms Track.
\wacvfinalcopy % *** Uncomment this line for the final submission

\ifwacvfinal
\usepackage[breaklinks=true,bookmarks=false]{hyperref}
\else
\usepackage[pagebackref=true,breaklinks=true,colorlinks,bookmarks=false]{hyperref}
\fi

\begin{document}

%%%%%%%%% TITLE
\title{Fine-grained Affordance Annotation for Egocentric Hand-Object Interaction Videos}

\author{Zecheng Yu, Yifei Huang, Ryosuke Furuta, Takuma Yagi, Yusuke Goutsu, Yoichi Sato\\
Industrial Institute of Science, The University of Tokyo\\
{\tt\small \{zch-yu,hyf,furuta,tyagi,goutsu,ysato\}@iis.u-tokyo.ac.jp}
}

\maketitle
\thispagestyle{empty}

%%%%%%%%% ABSTRACT
\begin{abstract}
   Object affordance is an important concept in hand-object interaction, providing information on action possibilities based on human motor capacity and objects' physical property thus benefiting tasks such as action anticipation and robot imitation learning. However, the definition of affordance in existing datasets often: 1) mix up affordance with object functionality; 2) confuse affordance with goal-related action; and 3) ignore human motor capacity. This paper proposes an efficient annotation scheme to address these issues by combining goal-irrelevant motor actions and grasp types as affordance labels and introducing the concept of mechanical action to represent the action possibilities between two objects. We provide new annotations by applying this scheme to the EPIC-KITCHENS dataset and test our annotation with tasks such as affordance recognition, hand-object interaction hotspots prediction, and cross-domain evaluation of affordance. The results show that models trained with our annotation can distinguish affordance from other concepts, predict fine-grained interaction possibilities on objects, and generalize through different domains.
\end{abstract}

%%%%%%%%% BODY TEXT
\section{Introduction}
\label{sec:intro}

Affordance was first defined by James Gibson~\cite{gibson1977concept} as the possibilities of action that objects or environments offer. It is a non-declarative knowledge we have learned for automatically activating afforded responses on an object decided by both our motor capacity, \textit{i.e.}, the motor actions suitable for human hands, and the object's physical properties such as shape. Recognizing affordance can benefit tasks like action anticipation and robot action planning by providing information on possible interactions with objects in the scene~\cite{liu2020forecasting}.

Many existing works~\cite{koppula2013learning, myers2015affordance, luddecke2017learning, nguyen2017object, thermos2017deep, nagarajan2019grounded} in computer vision investigated affordance. They use verbs as affordance labels to describe the possible actions associated with objects.
However, verbs like ``cut", ``take", and ``turn off" do not correspond the definition of affordance. More specifically: a) ``cut" is a possible action enabled by a knife, thus directly using ``cut" as an affordance label fails to distinguish human natural motor capacities from the capacities extended by objects' functionalities; 
b) using ``take" as an affordance label overlooks changes in affordance when ``take" is performed with different grasp types, which cannot provide fine-grained affordance annotations;
and c) ``turn-off" is a goal-related action, but not a goal-irrelevant affordance. The affordance utilized in ``turn-off tap" should also apply in other interactions such as ``press button". Instead of the confusion of affordance and other concepts, verbs also can't represent the diversity of affordance. For example, we cannot tell the differences between ``pick-up bowl" and ``pick-up knife" by the verb ``pick-up".

\begin{figure}[t]
    \centering
    \includegraphics[width=1.0\linewidth]{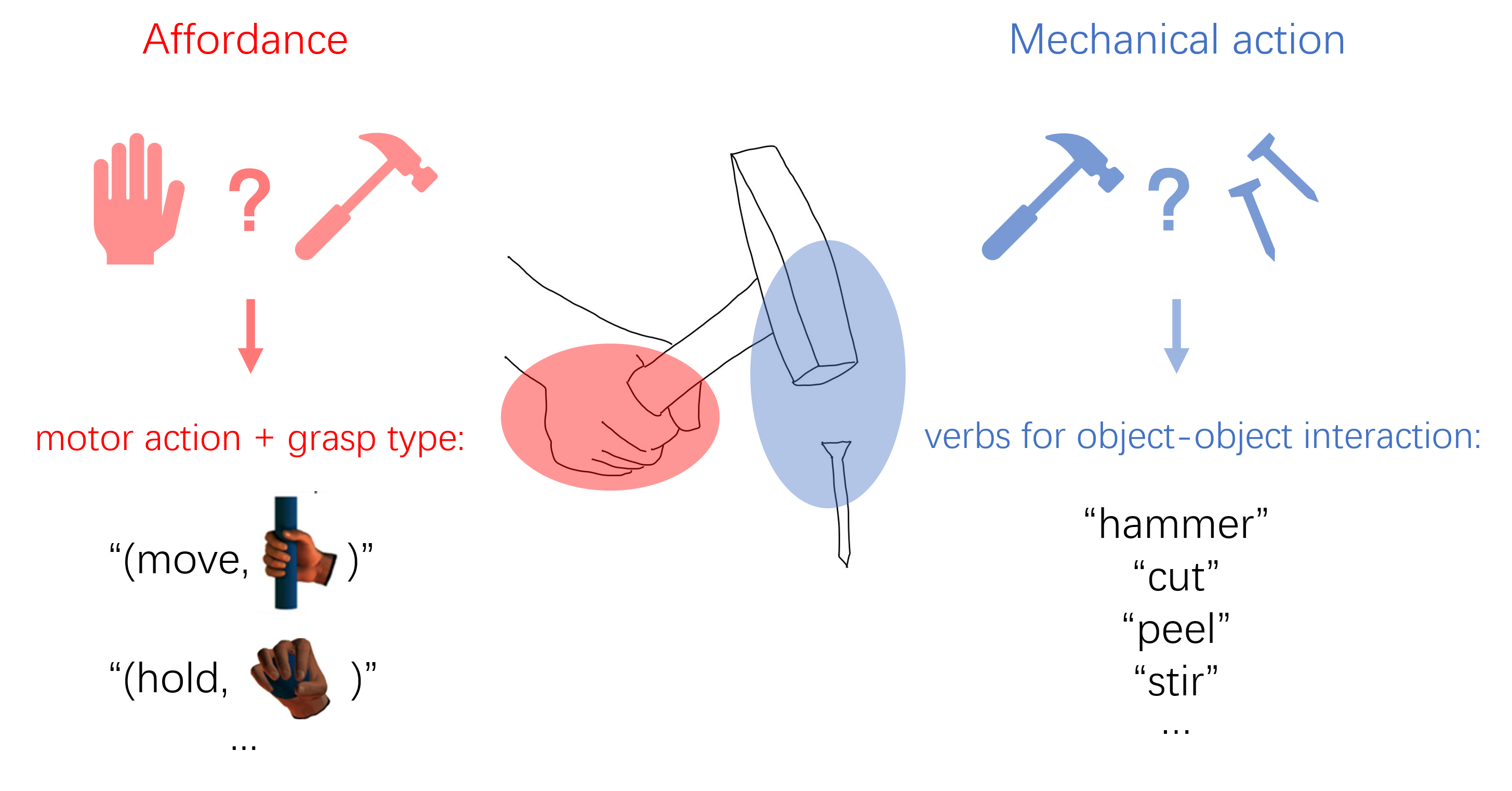}
    \caption{\textbf{Definitions used in our annotation scheme}: a) affordance labels are the combination of goal-irrelevant motor actions and grasp types that exists in hand-object interface; and b) mechanical action labels are action labels that can describe the interaction between two objects, exists in tool-object interface (figures adapted from ~\cite{osiurak2017affordance}).}
    %\ecaption{(a)Tool-Use Action. (b)Non-Tool-Use Action.}
    \label{fig:intro}
\end{figure}

In light of these issues, we need a precise affordance definition to distinguish affordance from other concepts, with the capability of representing human motor capacity. Inspired by the findings of neuroscience~\cite{osiurak2017affordance}, we address the shortcomings mentioned above by proposing an affordance annotation that considers hand-object interaction and tool-object interaction separately. Our annotation scheme: a) defines affordance as a combination of goal-irrelevant motor actions and hand grasp labels. This can represent the possible motor actions enabled by human motor capacity and object's physical property. This label structure can also provide a fine-grained affordance category to represent the diverse affordances in hand-object interactions; b) defines mechanical actions as the possible interactions between two objects, as shown in Figure~\ref{fig:intro}. 

Since annotating this information for a large-scale video dataset can be laborious, we propose an annotation method that leverages the consistency of affordance to simplify the annotation: the affordance will be the same when the same participant performs the same action on the same object. We apply this annotation scheme on the EPIC-KITCHENS dataset~\cite{damen2022rescaling}. The annotations are available at \url{https://github.com/zch-yu/epic-affordance-annotation}.

We test our annotation on three tasks to evaluate its rationality. First we compare the recognition results of affordance, mechanical action, and tool / non-tool use action, the results show that models trained with our annotation clearly distinguish affordance from other concepts. Then we generate hand-object interaction hotspots by using affordance as weakly-supervision followed ~\cite{nagarajan2019grounded}'s method, our results show that the model can predict a more fine-grained, accurate interaction hotspots compared with using action labels. Finally we evaluate the generalization capacity of affordance, the performance of affordance recognition on unseen domains is much better than action recognition.
%We believe that our newly proposed definition of affordance and the annotated dataset can facilitate a deeper understanding of object affordance and further improve subsequent tasks such as active object prediction, action anticipation, and robot imitation learning.

The main contributions of our work are as follows:
\begin{enumerate}
%\item We propose an precise definition for the annotation of affordance and other concepts related in human-object interaction field.
\item We point out the major shortcomings of existing affordance datasets, \textit{i.e.}, affordance is wrongly confused with object functionalities and goal-related actions. In addition, verbs cannot completely describe affordance because of neglecting the grasp types.
\item We propose a fine-grained and efficient affordance annotation scheme for hand-object interaction videos to address the issues above, and provide annotations of affordance and other related concepts for a large-scale egocentric action video dataset: EPIC-KITCHENS.
\item We test our annotation on tasks such as affordance recognition, hand-object interaction hotspots prediction, and cross-domain evaluation of affordance. The results show that models trained with our annotation can distinguish affordance from other concepts, predict fine-grained interaction possibilities on objects, and generalize through different domains.
\end{enumerate}

%-------------------------------------------------------------------------
\section{Related Works}
\label{sec:related}

\subsection{Affordance Datasets}

Earlier affordance datasets~\cite{nguyen2017object, myers2015affordance} annotated possible actions and the exact regions where actions could occur for object images. %Unlike video datasets that contain human-object interaction, image datasets fall short of providing large-scale interactive information. 
Koppula \textit{et al.}~\cite{koppula2013learning} provide affordance label annotation for human-object interaction video clips. Thermos \textit{et al.}~\cite{thermos2017deep} and Fang \textit{et al.}\cite{fang2018demo2vec} annotate human-object interaction hotspot maps as object affordance for video clips associated with their action labels. Furthermore, Nagarajan \textit{et al.}~\cite{nagarajan2019grounded} use the action labels of the EPIC-KITCHENS dataset, which is egocentric, as weak supervision to learn to generate human-object interaction hotspot maps. As we can see in Table~\ref{affordane datasets comparison}, these datasets all use verb as affordance label. They neither provide a clear definition of affordance nor consider humans' motor capacity. And they also fail to represent the diversity of affordance. Therefore, we propose a fine-grained affordance annotation scheme considering both humans' motor capacity and the object's physical property for hand-object interaction videos.

\begin{table*}
    \resizebox{\textwidth}{16mm}{
    \label{affordane datasets comparison}
    \centering
      \begin{tabular}{lcccccc} \toprule
         & Format & Categories & Image/Video & Interaction Region & View & Affordance Labels\\ \midrule
        IIT-AFF~\cite{nguyen2017object} & RGB Image & 9 & 8,835 & $\surd$ & - & \makecell{contain, cut, display, engine, grasp, \\ hit, pound, support, w-grasp} \\
        RGB-D Part Affordance Dataset~\cite{myers2015affordance} & RGB-D Image & 7 & 105 & $\surd$ & - & \makecell{grasp, cut, scoop, contain, \\ pound, support, warp-grasp} \\
        CAD-120~\cite{koppula2013learning} & RGB Video & 6 & 130 & $\times$ & third-person view & \makecell{openable, cuttable, pourable, containable, supportable, holdable} \\
        SOR3D~\cite{thermos2017deep} & RGB-D Video & 13 & 9,683 & $\surd$ & third-person view & \makecell{grasp, lift, push, rotate, open, hammer, cut, \\ pour, squeeze, unlock, paint, write, type} \\
        OPRA~\cite{fang2018demo2vec} & RGB Video & 7 & 11,505 & $\surd$ & third-person view & \makecell{hold, touch, rotate, push, pull, pick up, put down} \\
        EPIC-KITCHENS Affordance~\cite{nagarajan2019grounded} & RGB Image & 20 & 1,871 & $\surd$ & egocentric & subset of EPIC-KITCHENS action set \\ \bottomrule
      \end{tabular}}
    \caption{Comparative overview of existing affordance datasets}
\end{table*}

\subsection{Affordance Understanding}

Affordance understanding methods can be divided into four categories: Affordance Recognition, Affordance Semantic Segmentation, Interaction Hotspots Prediction, and Affordance as Context. Given a set of images/videos, the task of affordance recognition~\cite{azuma_estimation_nodate} aims to estimate affordance labels from them. Affordance semantic segmentation~\cite{luddecke2017learning, nguyen2017object} aims at segmenting the input image / video frame into a set of regions that are labeled with affordance labels. Interaction hotspots prediction~\cite{nagarajan2019grounded, fang2018demo2vec} tries to predict the possible interaction hotspots of objects. Moreover, some works ~\cite{koppula2015anticipating,liu2020forecasting,nagarajan2020ego} also use affordance as a context for other tasks such as action anticipation.

All of these methods are influenced by simply using verbs as affordance labels. Firstly, verbs confuse affordance with other concepts such as object functionality. For example, our attention differs when observing the ``cut'' action and the ``take'' action~\cite{huang2018predicting,huang2020mutual}: the former on the interacting object and the latter on the hand. Mixing them up may confuse models for affordance recognition. And verbs cannot represent the diverse affordances utilized in hand-object interaction. We may perform the same action with different affordances depending on the objects. For example, we may directly push or handle the doorknob to close a door. However, previous works overlook these details by simply using the action "close" as an affordance label. This leads to the failure to distinguish different affordance regions in affordance semantic segmentation and interaction hotspots prediction tasks.

\section{Proposed Affordance Annotation}
\label{sec:method}

Our goal is to develop a fine-grained and efficient annotation scheme for affordance and other related concepts for hand-object interaction videos.

\subsection{Definitions}

Our proposed affordance annotation scheme is inspired by the three-action system model (3AS)~\cite{osiurak2017affordance}, which clearly defines the concepts needed in hand-object interactions. The three-action system model includes affordance, mechanical action, and contextual relationship. We mainly focus on the first two concepts since they are closely related to our goal. Affordance is thus defined as
\begin{itemize}
    \item \textbf{Hand-centered:} affordance only presence in the hand-object interface.
    \item \textbf{Animal-relative:} affordance is not only determined by the properties of objects but is also related to human motor capacity.
    \item \textbf{Goal-irrelevant object property:} the same affordance can be utilized for different purposes.
\end{itemize}
To fill the absence of affordance in object-object interactions, the 3AS~\cite{osiurak2017affordance} introduce mechanical actions as tool-centered, mechanical action possibilities between objects. According to the above, we separately consider the affordance and mechanical action for an instance of hand-object interaction as shown in Figure~\ref{fig:intro}. We define affordance label as a combination of goal-irrelevant motor actions and grasp types. For mechanical actions, we use verbs that describe interactions between objects (such as \textit{cut, stir}) as mechanical action labels.

\begin{figure}[t]
    \centering
    \includegraphics[width=1.0\linewidth]{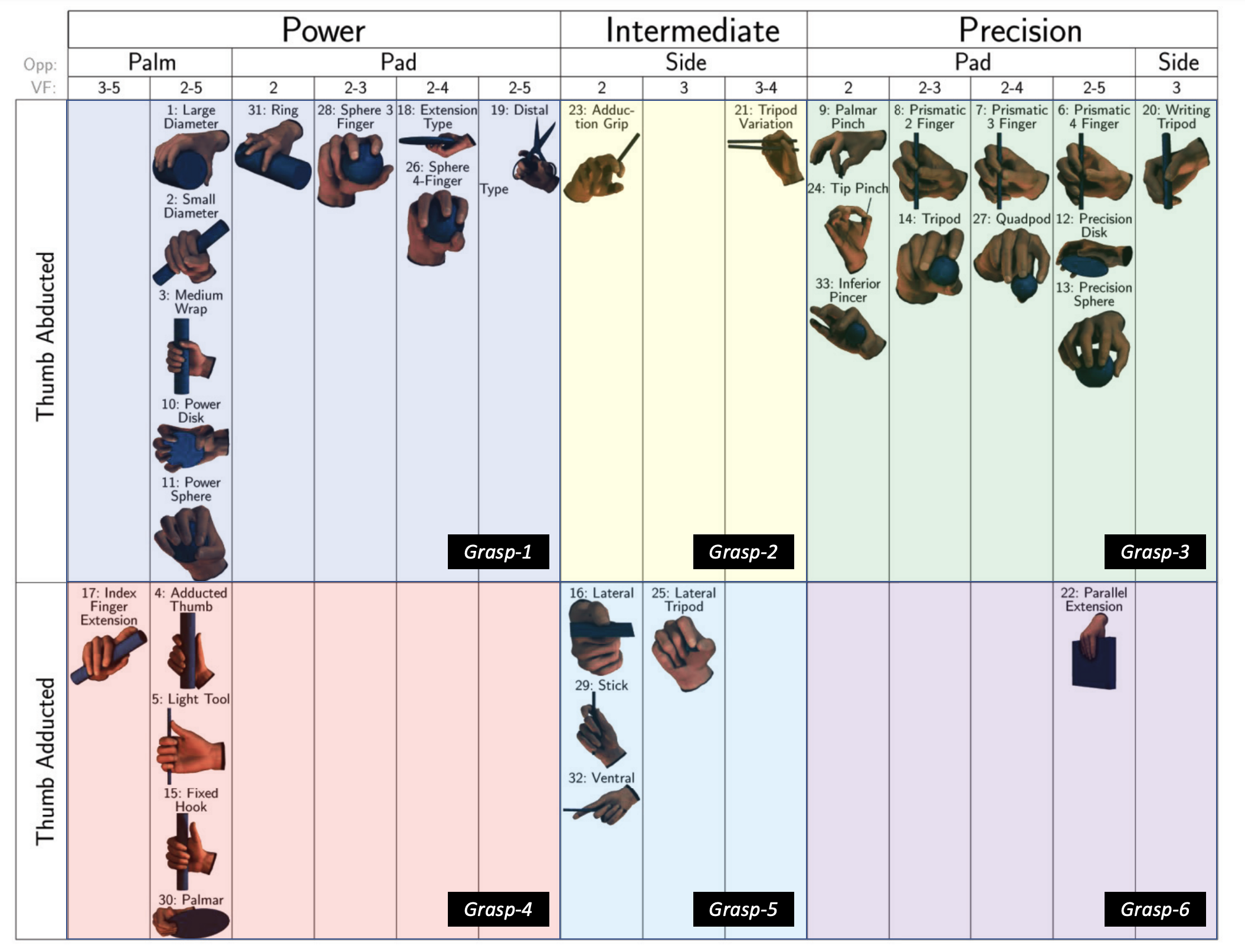}
    \caption{The 6-class grasp type taxonomy simplified from~\cite{feix2015grasp}.}
    \label{fig:grasptype}
\end{figure}

\begin{figure}[t]
    \centering
    \includegraphics[width=1.0\linewidth]{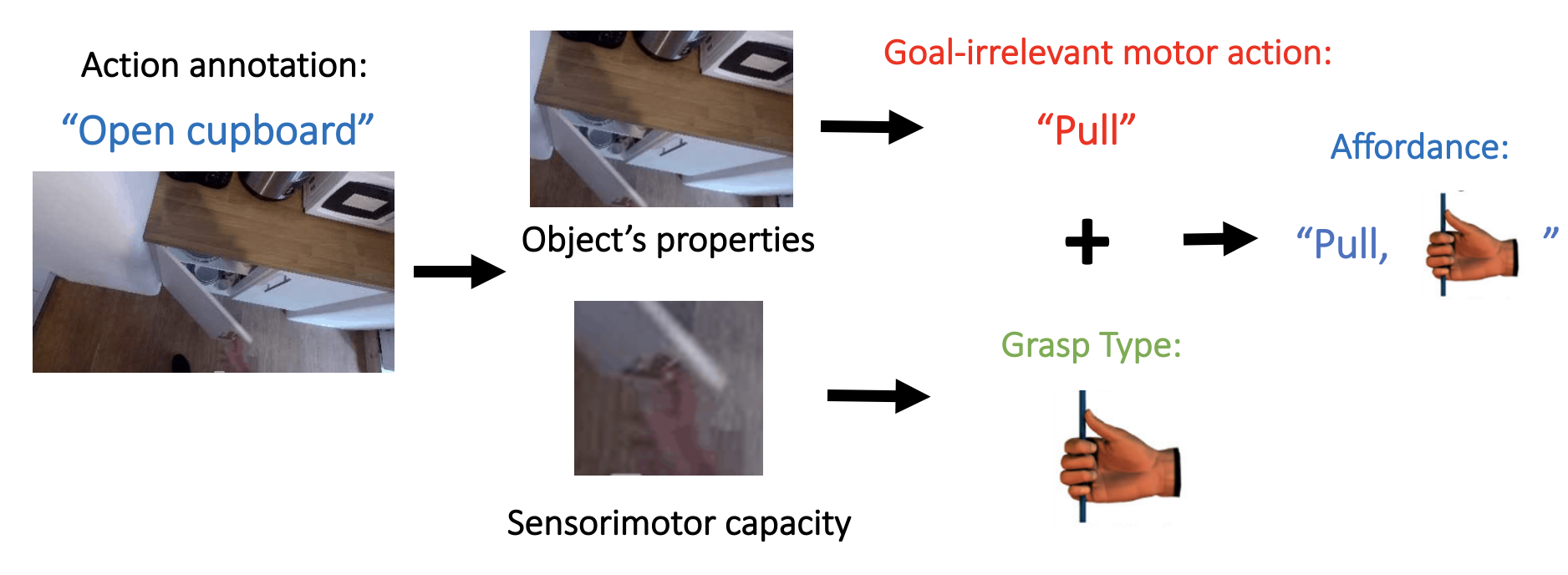}
    \caption{An affordance label is composed of a goal-irrelevant motor action label and a grasp type label.}
    \label{fig:affordancelabel}
\end{figure}

\subsection{Annotation Scheme}

To annotate an action video dataset, we first need to divide the original action labels of the dataset into tool-use actions and non-tool-use actions since mechanical actions are only present in tool-use actions. We then annotate mechanical actions for tool-use actions and affordances for both tool-use and non-tool-use actions. Directly annotating these labels can be laborious, thus we proposed an annotation scheme which utilizes existing annotations to reduce the difficulty of labeling.

\begin{figure*}[t]
    \centering
    \includegraphics[width=1.0\linewidth]{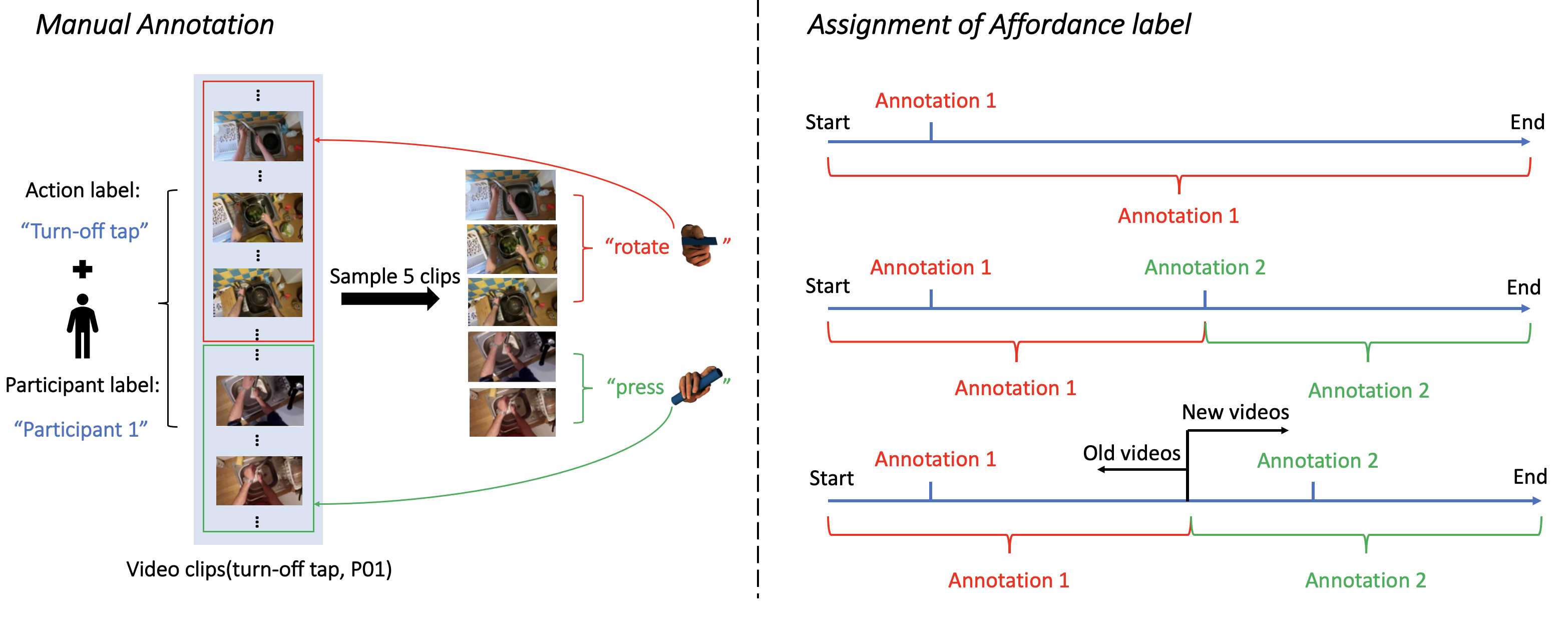}
    \caption{\textbf{The efficient annotation scheme.}
    \textbf{Left}: We first sample 5 video clips from the clips annotated with the same action-participant label, then manually annotate affordance label for them. \textbf{Right}: after manual annotation, the affordance labels are assigned to all video clips with the same action-participant label as follows: (a) If there is only one affordance label, then it is assigned to all videos; (b) If there are two affordance labels, we assign annotation 2 to video clips later than the video clip which is annotated with annotation 2, annotation 1 to video clips former than it; (c) If there is already a boundary which divides the video clips into groups, videos in each group are annotated with the annotation located in their group.}
    \label{fig:efficient}
\end{figure*}

\textbf{Tool-use / non-tool-use action annotation}: Tool-use / non-tool-use action annotation for action video datasets can be done by dividing original action labels of the dataset into three categories: tool-use action, non-tool-use action, and both, according to the meaning of each action label. For example, ``take” is a non-tool-use action, while ``cut" is a tool-use action. Some action labels could represent both the tool-use action and the non-tool-use action simultaneously, such as ``wash''. We ignore these labels during annotation because of their ambiguity.

\textbf{Mechanical action annotation}: We only need to annotate mechanical actions for tool-use actions. According to the definition, we can use the verbs in original action labels as mechanical action labels. For example: in ``stir food'', ``stir'' can represent the mechanical action between the slice and the food. We can automatically annotate mechanical actions for all tool-use action video clips based on this rule, allowing significant reduction of annotation cost.

\textbf{Affordance annotation}: For affordance annotation, as shown in Figure~\ref{fig:affordancelabel}, we annotate a goal-irrelevant motor action and a grasp type for each video clip. Given an unlabeled video clip, we first define a goal-irrelevant motor action according to the object property used in it. In the example of this figure, we use ``pull'' to represent the ``pullable'' property of the cupboard's handle. Next, we chose a grasp type from a 6-class grasp type taxonomy. This taxonomy is simplified from a well-known 33-class grasp type taxonomy~\cite{feix2015grasp} based on the power of the grasp type and the posture of the thumb, as shown in Figure~\ref{fig:grasptype}. Finally, we combine the goal-irrelevant motor action label with the grasp-type label as the affordance label. This form of affordance labels can model both the object's physical property and the human motor capacity. And the combination of motor action and grasp type provides a fine-grained structure to represent diverse affordances.

%Besides
To reduce the manual work of affordance annotation, we propose an efficient annotation method based on the assumption that the same person would interact with the same object in a fixed manner. As shown Figure~\ref{fig:efficient}(left), there are multiple video clips demonstrating the same participant performs the same action in the original dataset. We first sample five clips from video clips with the same action-participant (verb, noun, participant) annotation, then manually annotate affordances for them with the CVAT~\cite{boris_sekachev_2020_4009388}. Then we assign these affordance labels to video clips with the same action-participant annotation, as shown in Figure~\ref{fig:efficient}. In some cases, videos with the same action-participant label have multiple affordance labels because of the scene variation (\textit{e.g.}, the participant performs the same action in different rooms). To address this issue, our affordance annotation assignment scheme is as follows: (a) One affordance annotation: if there is no scene change among these video clips, we apply the only affordance annotation to all video clips with the same action-participant label. (b) Two scenes without a predefined boundary (\textit{e.g.}, the boundary of EPIC-KITCHENS55~\cite{Damen2018EPICKITCHENS} and EPIC-KITCHENS100): we use affordance occurred later as a boundary, video clips earlier than it are annotated with annotation 1, those later than it are annotated with annotation 2. (c) Two scenes with a boundary: There exists a boundary which divides the video clips into two groups based on their relative position to the boundary. Videos of each group are annotated with the annotation inside their group. Note that there is a trade-off between annotation speed and accuracy. The more video clips we manually annotate the more accurate the automatic annotation results will be. The efficiency and accuracy of this method are shown in Section~\ref{sec:data}.

\begin{figure*}[t]
    \centering
    \includegraphics[width=1.0\linewidth]{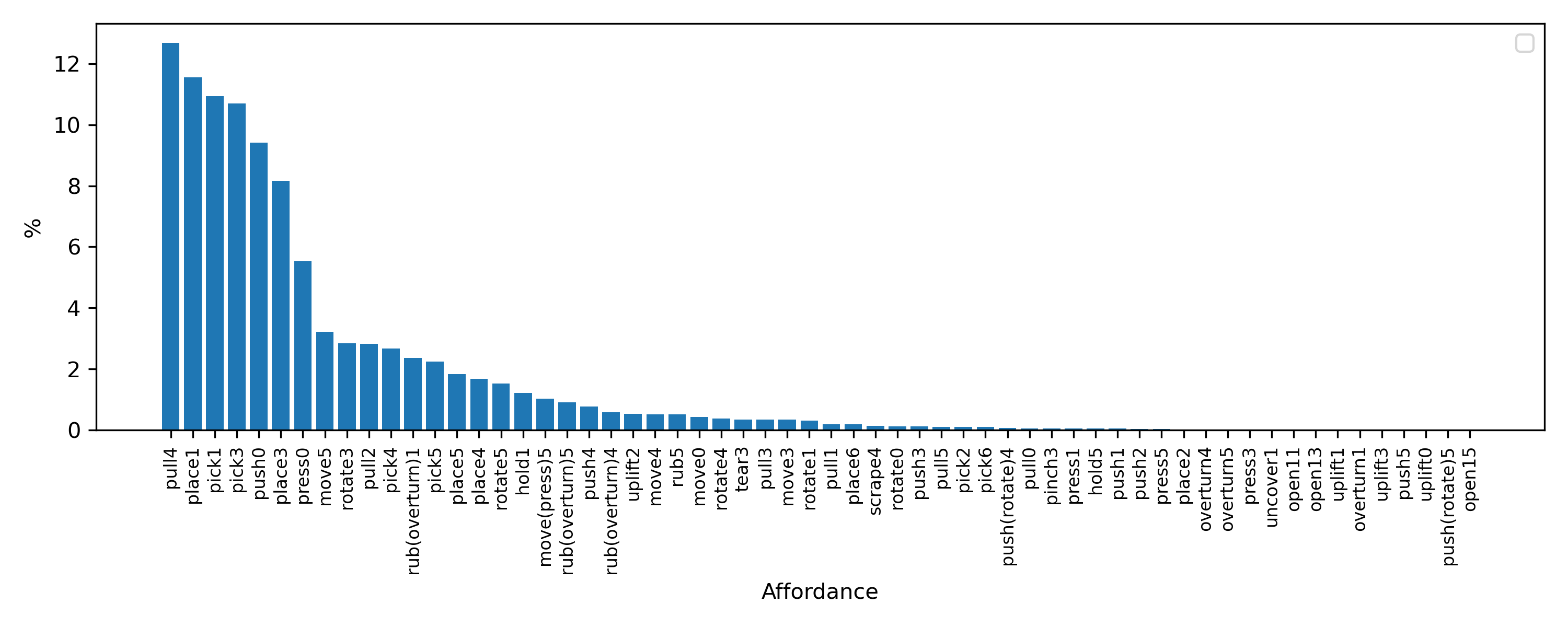}
    \caption{Distribution of affordance classes}
    \label{fig:affordance_dis}
\end{figure*}

\section{Experiments}

We first annotate the EPIC-KITCHENS dataset with our proposed method and then train models on them. Our experiments evaluate the rationality of the annotation on four aspects: first, we test the recognition performance of tool use action / non-tool use action to demonstrate the importance of separating these action domains. Secondly, we compare the recognition models of affordance and mechanical action, evaluating the models' capacity of distinguishing affordance from other concepts. Thirdly, we show their capability of representing the diverse affordances by following Nagarajan \textit{et al.}~\cite{nagarajan2019grounded}'s method. Finally, we compare the generalization ability of affordance and action.

\subsection{Dataset}
\label{sec:data}
\begin{figure}[t]
    \centering
    \includegraphics[width=1.0\linewidth]{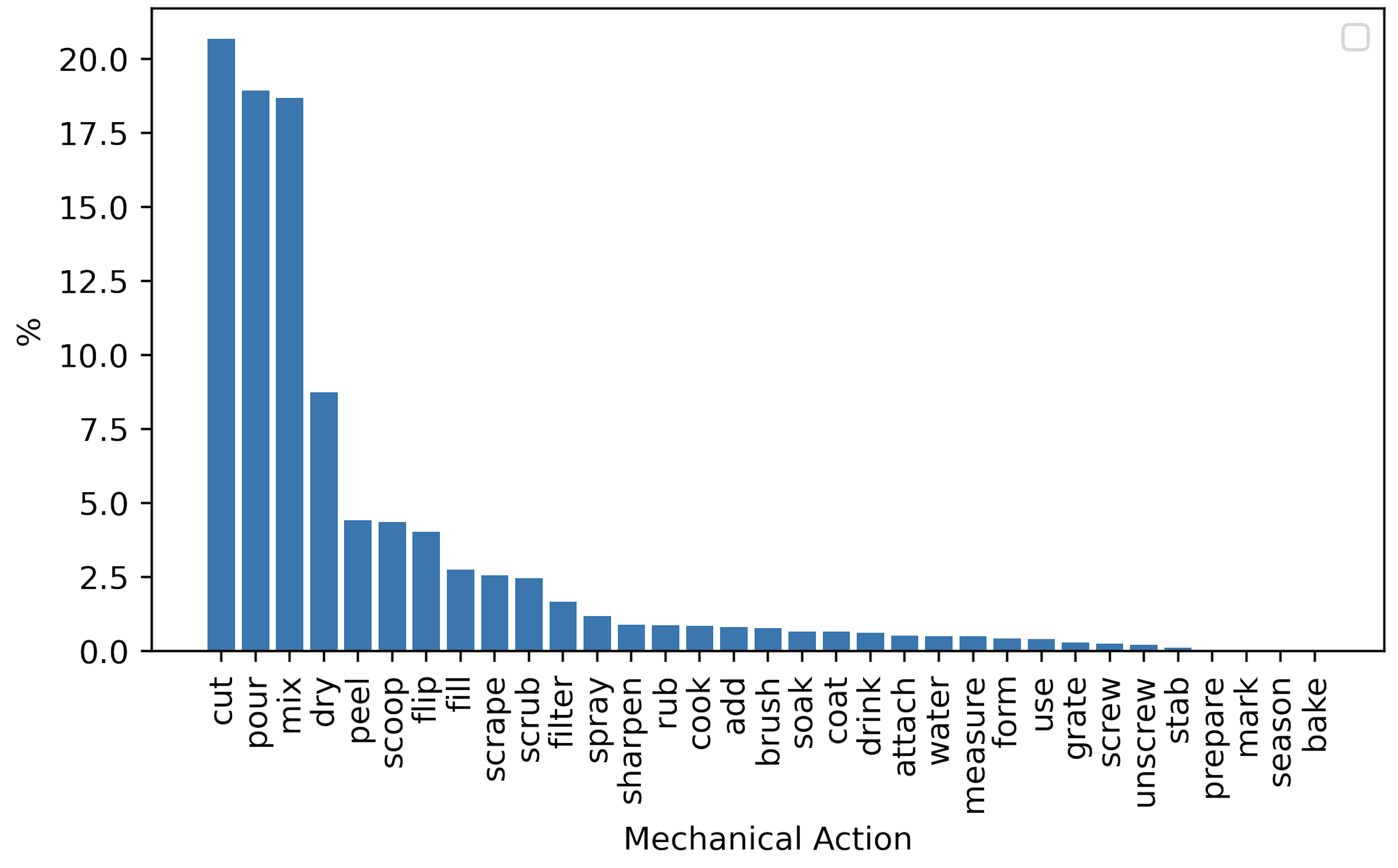}
    \caption{Distribution of mechanical classes}
    \label{fig:mechanical_dis}
\end{figure}

We chose the EPIC-KITCHENS dataset among large scale video datasets~\cite{damen2022rescaling,carreira2019short,goyal2017something}, which contains egocentric hand-object interaction video clips annotated with action (verb, noun) labels, recorded by 21 participants in 45 kitchen scenes. We first annotate tool use action / non-tool use action for the dataset. There are 60 non-tool use actions and 33 tool use actions among the 97 verbs of EPIC-KITCHENS,  which results in 51.5k non-tool use action video clips and 8.5k tool use action video clips, as shown in Table~\ref{epickitchenstoolnontool}. Then the verb annotation of the tool use action video clips are used as their mechanical action labels, as shown in Figure~\ref{fig:mechanical_dis}. Next, using our annotation method, we sampled 91 most used action labels from the original annotation and manually annotated 1356 affordance labels for different action-participant pairs. We obtain 18, 613 video clips annotated with 60 affordance labels, as shown in Figure~\ref{fig:affordance_dis}. For quality check, we randomly sample 1,113 instances from the annotated video clips and manually checked their affordance label, getting an accuracy of 87.95\%, and then correct the affordance annotation with them.

\begin{table}[t]
\begin{center}
      \begin{tabular}{lcc}
        \toprule
        Category & Video Clips & Action labels \\ 
        \midrule
        Non-tool use & 51.5k & take, put, open...(60) \\
        Tool use & 8.5k & cut, pour, mix, dry...(33) \\
        \bottomrule
      \end{tabular}
\end{center}
\caption{Tool-use / non-tool-use action labels of the EPIC-KITCHENS}
\label{epickitchenstoolnontool}
\end{table}

\subsection{Identification of Affordance / Mechanical Action / Tool Use Action}
\label{recognition}
 
To evaluate whether our annotation scheme can address the issue of confusing affordance with object functionality and goal-related actions. We train three recognition models for tool / non-tool use action, mechanical action, and affordance separately, and compare their visualization results to show the important regions each model focuses on.

\textbf{Tool Use / Non-Tool Use Action Recognition}: We train two SlowFast~\cite{feichtenhofer2019slowfast} models to recognize tool use action / non-tool use action from a given video clip, one with random tool / non-tool use action annotation and another with our annotation. The results are shown in Table~\ref{tabletoolnontool}, which demonstrates the rationality of our automatic annotation scheme for tool / non-tool use actions.

\textbf{Mechanical Action Recognition}: We train a SlowFast model for mechanical action recognition with our annotation. For the 33-class mechanical action recognition task, as shown in Table~\ref{affordanceresult}, we get a recognition accuracy of 51.90\% .

\textbf{Affordance Recognition}: We train a SlowFast model for affordance recognition with our annotation. For the 60-class affordance recognition task, as shown in Table~\ref{affordanceresult}, we get a recognition accuracy of 57.08\%.

\begin{table}[t]
\begin{center}
  \scalebox{0.9}{
  \begin{tabular}{lcc}
    \toprule
    Dataset & Tool use actions & Non tool-use actions \\
    \midrule
    Random annotation & 0.4720 & 0.5282 \\ 
    Our annotation & 0.8580 & 0.7867\\ 
    \bottomrule
  \end{tabular}
  }
\end{center}
\caption{Tool-use / non-tool use action classification results.}
\label{tabletoolnontool}
\end{table}

\begin{figure}[t]
    \centering
    \includegraphics[width=1.0\linewidth]{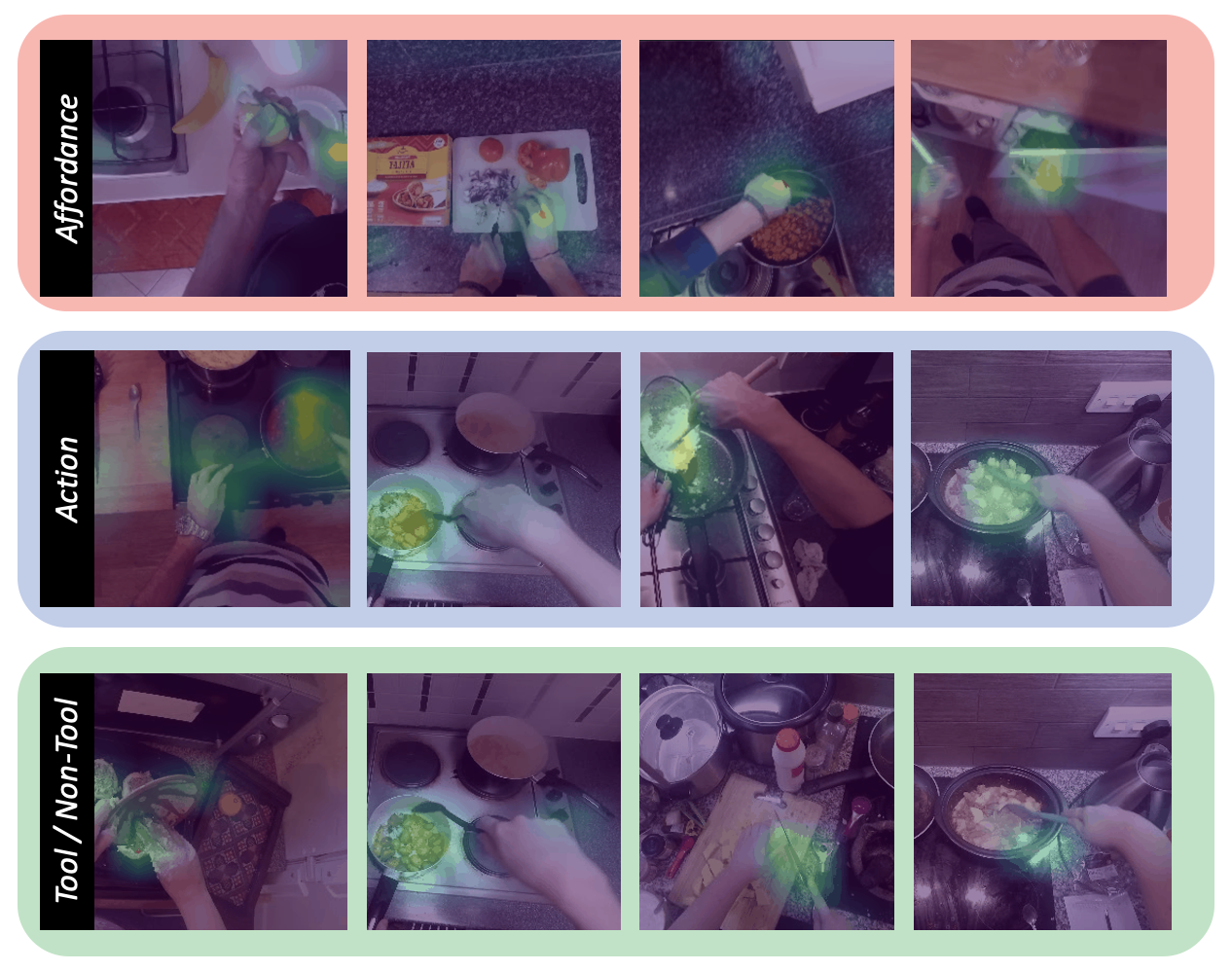}
    \caption{\textbf{GradCam~\cite{selvaraju2017grad} visualization results of affordance recognition, mechanical action recognition, and tool-use/non-tool-use action recognition.} The affordance recognition model focuses on the hand-object interaction, while the mechanical action recognition model cares more about object-object interaction. This demonstrate our proposed annotation labels clearly separate affordance with other concepts, and completely represent affordance without missing the human motor capacity.}
    \label{fig:comparison}
\end{figure}

According to the three-action model, affordances are only present in the hand-object interface, and mechanical actions are only present between the object-object interface. Thus, interaction regions can help us distinguish affordance from object functionality. Here, we compare the visualization results of these models to see whether our proposed annotation scheme can separate affordance with other concepts. The visualization results generated by GradCam~\cite{selvaraju2017grad} are shown in Figure ~\ref{fig:comparison}. From the first row, we can see that the affordance recognition model focuses more on the hand-object interaction. The second row shows that the mechanical action recognition model cares more about object interaction. When the second row is compared to the third row, it is evident that the mechanical action recognition model focuses on tool-object interactions (2nd row), and the tool use/non-tool use action classification model focuses on the existence of tools (3rd row). These results show that our affordance annotation provides a more precise affordance label than previous datasets, which is important for discriminating affordance with other concepts.    

\begin{table}[t]
\begin{center}
  \scalebox{1.0}{
  \begin{tabular}{lcc}
    \toprule
    & Top1 Acc & Top5 Acc  \\
    \midrule
    Affordance& 0.5708 & 0.8771 \\ 
    Mechanical action& 0.5190 & 0.8643 \\
    \bottomrule
  \end{tabular}
  }
  
\end{center}
\caption{Affordance / mechanical action recognition results}
\label{affordanceresult}
\end{table}

\subsection{Hand-Object Interaction Hotspots}
\label{hotspots}

In this section, we evaluate our affordance annotation on the interaction hotspot prediction task, which generates an interaction hotspot map for a given object image. We follow the weakly-supervised method proposed in ~\cite{nagarajan2019grounded}. They first train an action recognition model with action labels and an anticipation network that can predict the ``active state" of an inactive object. At the inference phase, the input image is fed to the anticipation network, followed by the action recognition model to obtain the action prediction. Then they derive gradient-weighted attention maps as the interaction hotspot map. In our experiment, we train the model with our affordance labels instead of action labels.

We first compare the ground-truth heatmaps to our affordance-supervised interaction hotspots and the action-supervised interaction hotspots. We train the affordance model with 4, 344 video clips including 43 different affordance labels, and the action model with 9, 236 video clips including 20 different action labels. Table~\ref{hotspotsresults} shows the results, we report the error as KL-Divergence ~\cite{selvaraju2017grad}, SIM, and AUC-J ~\cite{bylinskii2018different}. The affordance model outperforms the action model in these metrics, demonstrating that the model can better capture the interaction cues with the supervision of affordance labels. One reason is that the affordance label's components: goal-irrelevant action and grasp type are more suitable for representing hand-object interactions. Furthermore, the granularity of affordance labels also helps the model avoid missing possible interactions.

\begin{figure}[t]
    \centering
    \includegraphics[width=1.0\linewidth]{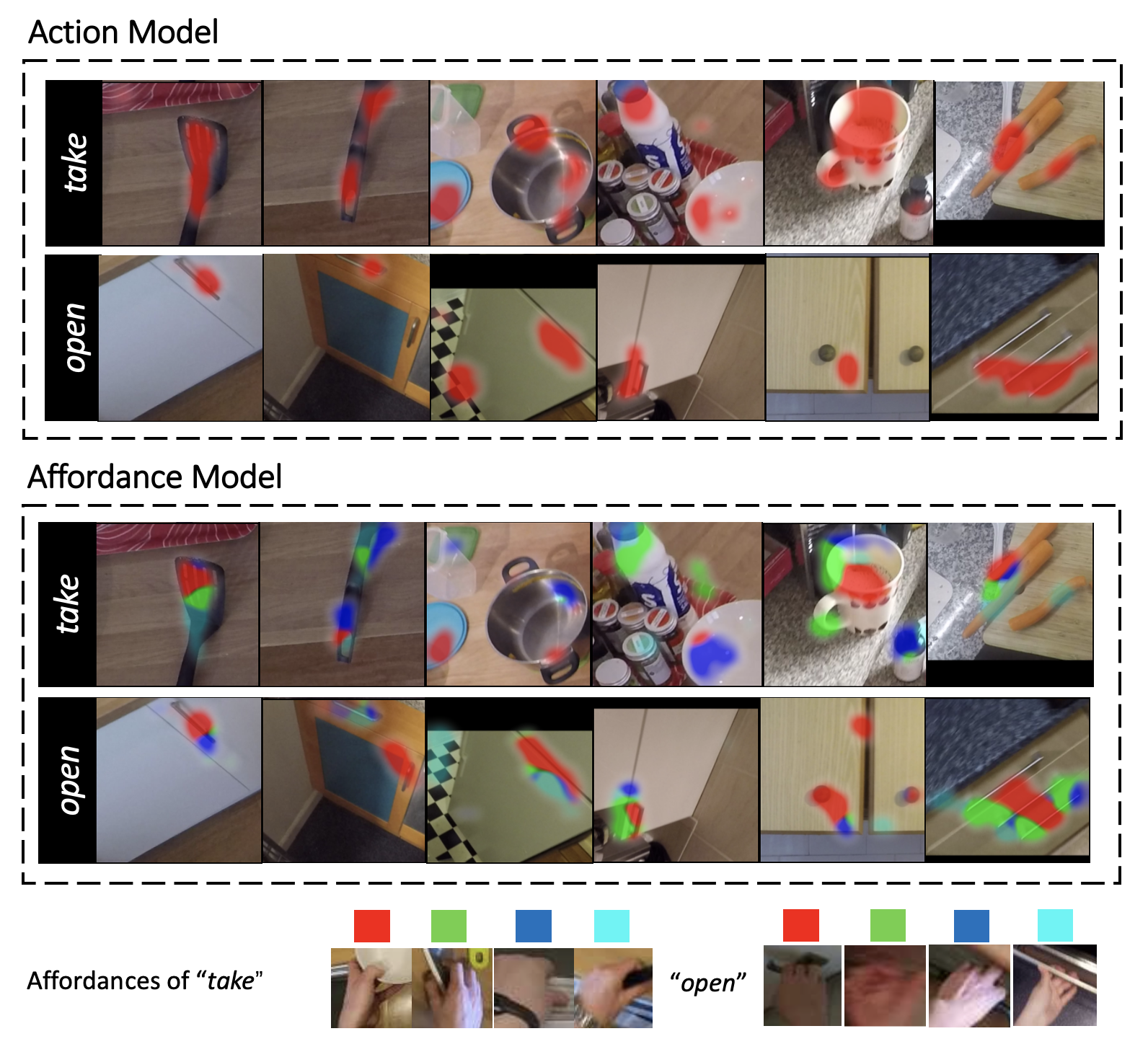}
    \caption{\textbf{Interaction heatmaps generated by action models and affordance models.} \textbf{Top rows:} Interaction heatmaps of \textit{take, open}(red) generated using action model. \textbf{Bottom rows:} Interaction heatmaps of affordances(red, green, blue, cyan) used in \textit{take, open} generated using affordance model. The affordance model trained with our annotation performs better on both capturing the right interaction region and representing the diversity of hand-object interaction.}
    \label{fig:hotspotexample}
\end{figure}

In addition to the quantitative performance improvement, the predicted interaction heatmaps also benefit from the fine-grained affordance labels of our annotation. The interaction heatmaps generated by the action model and affordance model are shown in Figure~\ref{fig:hotspotexample}, which highlight the interaction regions of different actions(affordances) on objects. The top rows show that the action model can predict the possible interaction regions on different objects(\textit{e.g.}, we can take the slice by grasping the red regions in the first image). Comparing these heatmaps with those in the bottom rows generated by the affordance model, we can see that the affordance heatmaps can better represent the diverse hand-object interactions(\textit{e.g.}, the ``take" action are replaced by four different affordances, representing different interaction ways of ``taking"). This demonstrates our affordance annotation's capacity to help the model learn a more granular representation of hand-object interaction.

\begin{table}[t]
\begin{center}
\scalebox{1.0}{
  \begin{tabular}{lccc}
    \toprule
    & KLD$\downarrow$ & SIM$\uparrow$ & AUC-J$\uparrow$  \\
    \midrule
    Action label & 2.661 & 0.382 & 0.758 \\ 
    Affordance label & 1.305 & 0.399 & 0.776 \\ 
    \bottomrule
  \end{tabular}}
  
\end{center}
\caption{\textbf{Interaction hotspots prediction results generated using action model and affordance model.} The affordance model outperforms the action model on all metrics, demonstrating that our fine-grained affordance labels can help the model better capture the interaction cues.}
\label{hotspotsresults}
\end{table}

\iffalse
\begin{figure}[t]
    \centering
    \includegraphics[width=0.9\linewidth]{fig/hotspotsinterface.png}
    \caption{Interaction hotpots generated by affordance model can }
    \label{fig:hotspotinterface}
\end{figure}
\fi

\begin{table*}[t]
\begin{center}
  \begin{tabular}{lcccc}
    \toprule
    Task & Target Top-1 Acc & Target Top-5 Acc & Source Top-1 Acc & Source Top-5 Acc\\
    \midrule
    Action Recognition & 12.67 & 30.99 & 16.16 & 32.91\\ 
    Affordance Recognition & 20.69 & 54.59 & 24.68 & 53.58 \\ 
    \bottomrule
  \end{tabular}
  
\end{center}
\caption{\textbf{Cross-domain recognition results of action / affordance}:the affordance recognition models work better in the unseen target domain.}
\label{crossdomainresults}
\end{table*}

\subsection{Generalization Ability of Affordance on Different Domains}

In this section, we evaluate the generalization ability of the affordance recognition model. How well does the model trained on videos from one domain work on other domains? We followed the experiment setting of the unsupervised domain adaptation(UDA) challenge of the EPIC-KITCHENS dataset. First, we separate annotated video clips into source and target domains according to EPIC-KITCHENS's UDA setting. We then train an affordance recognition model and an action recognition model in the source domain. Finally, the model which works the best on the source domain's validation set is chosen to be evaluated on the target domain. Table~\ref{crossdomainresults} shows the comparison results of affordance recognition and action recognition. Here we use the same Slowfast model for both tasks. We can see that the performance drop of the affordance recognition model in the target domain is smaller than that of the action model, which demonstrates the generalization ability of the learned affordance representation.

\begin{figure}[t]
    \centering
    \includegraphics[width=1.0\linewidth]{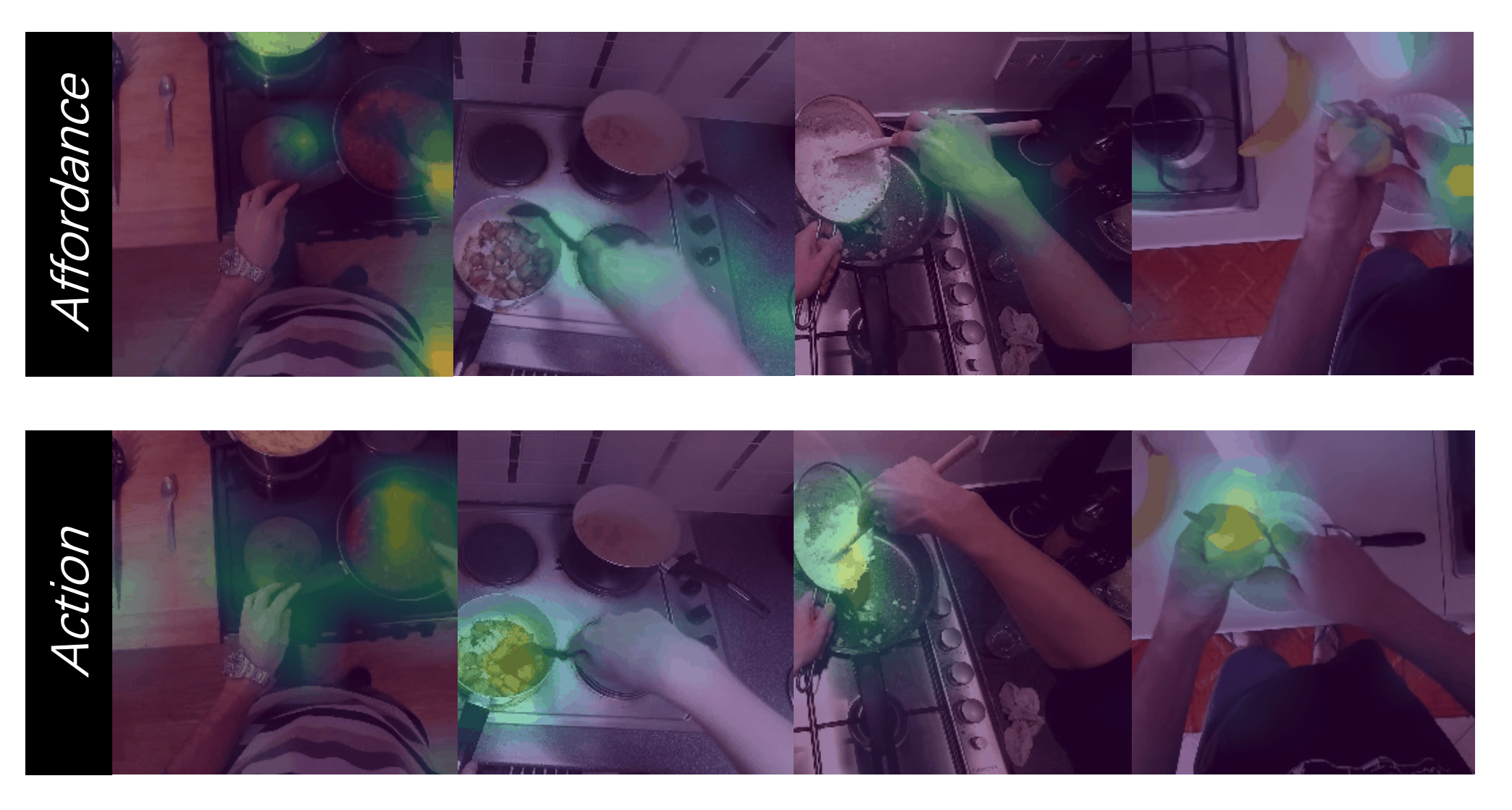}
    \caption{\textbf{Visualization results of the affordance recognition model(top) and the action recognition model(bottom).} Our affordance annotation guide the model to focus on the hand-object interaction which does not vary much with the domain. Thus, the learned affordance representation's generalization ability is much better than the action representation.}
    \label{fig:crossdomain}
\end{figure}

Figure~\ref{fig:crossdomain} shows the visualization results of the affordance recognition model(top), and the action recognition model(bottom). We can see that the affordance model focuses more on the hand-object interaction, but the action model cares more about the target object. The reason is that the affordance labels(\textit{goal-irrelevant action, grasp type}) focus more on how we interact with the object, such as the appearance and motion of the hand, and the parts of the object to be interacted with. These do not vary much with different domains. However, the action labels(\textit{verb, noun}) focus on the object and human-object interaction, which are difficult to be consistent through different domains. Objects with the same label may change a lot through domains(\textit{e.g.}, knives in a different color, size.). Furthermore, human-object interactions are also changeable since different people may perform the same action differently. Thus, the affordance recognition model works better in the target domain.

\section{Conclusion}

In this study, we proposed a fine-grained affordance annotation scheme for hand-object interaction videos, distinguishing affordance from other concepts like object functionality and reducing the manual annotation burden. We successfully applied our proposed annotation scheme to the EPIC-KITCHENS dataset and evaluated them on three tasks. The results of recognition tasks on affordance, mechanical action, and tool / non-tool use action show that our proposed annotation clearly distinguishes affordance from other concepts and completely represents affordance without missing human motor capacities. The interaction hotspots prediction results demonstrate that our fine-grained affordance labels can better represent the diverse hand-object interactions than verbs. Moreover, our affordance annotation also exhibits its generalization ability in different domains.
{\small
\section*{Acknowledgments}
This work is supported by JST AIP Acceleration Research Grant Number JPMJCR20U1 and JSPS KAKENHI Grant Numbers JP20H04205 and JP22K17905.
}
\clearpage
{\small
\bibliographystyle{ieee_fullname}
\bibliography{egbib}

\begin{thebibliography}{10}\itemsep=-1pt

\bibitem{azuma_estimation_nodate}
Ryunosuke Azuma, Tetsuya Takiguchi, and Yasuo Ariki.
\newblock Estimation of {Object} {Functions} {Using} {Visual} {Attention}.
\newblock page~4, 2018.

\bibitem{bylinskii2018different}
Zoya Bylinskii, Tilke Judd, Aude Oliva, Antonio Torralba, and Fr{\'e}do Durand.
\newblock What do different evaluation metrics tell us about saliency models?
\newblock {\em IEEE transactions on pattern analysis and machine intelligence},
  41(3):740--757, 2018.

\bibitem{carreira2019short}
Joao Carreira, Eric Noland, Chloe Hillier, and Andrew Zisserman.
\newblock A short note on the kinetics-700 human action dataset.
\newblock {\em arXiv preprint arXiv:1907.06987}, 2019.

\bibitem{Damen2018EPICKITCHENS}
Dima Damen, Hazel Doughty, Giovanni~Maria Farinella, Sanja Fidler, Antonino
  Furnari, Evangelos Kazakos, Davide Moltisanti, Jonathan Munro, Toby Perrett,
  Will Price, and Michael Wray.
\newblock Scaling egocentric vision: The epic-kitchens dataset.
\newblock In {\em European Conference on Computer Vision (ECCV)}, 2018.

\bibitem{damen2022rescaling}
Dima Damen, Hazel Doughty, Giovanni~Maria Farinella, Antonino Furnari,
  Evangelos Kazakos, Jian Ma, Davide Moltisanti, Jonathan Munro, Toby Perrett,
  Will Price, et~al.
\newblock Rescaling egocentric vision: collection, pipeline and challenges for
  epic-kitchens-100.
\newblock {\em International Journal of Computer Vision}, 130(1):33--55, 2022.

\bibitem{fang2018demo2vec}
Kuan Fang, Te-Lin Wu, Daniel Yang, Silvio Savarese, and Joseph~J Lim.
\newblock Demo2vec: Reasoning object affordances from online videos.
\newblock In {\em CVPR}, pages 2139--2147, 2018.

\bibitem{feichtenhofer2019slowfast}
Christoph Feichtenhofer, Haoqi Fan, Jitendra Malik, and Kaiming He.
\newblock Slowfast networks for video recognition.
\newblock In {\em ICCV}, pages 6202--6211, 2019.

\bibitem{feix2015grasp}
Thomas Feix, Javier Romero, Heinz-Bodo Schmiedmayer, Aaron~M Dollar, and Danica
  Kragic.
\newblock The grasp taxonomy of human grasp types.
\newblock {\em IEEE Transactions on human-machine systems}, 46(1):66--77, 2015.

\bibitem{gibson1977concept}
James~J Gibson.
\newblock The concept of affordances.
\newblock {\em Perceiving, acting, and knowing}, 1, 1977.

\bibitem{goyal2017something}
Raghav Goyal, Samira Ebrahimi~Kahou, Vincent Michalski, Joanna Materzynska,
  Susanne Westphal, Heuna Kim, Valentin Haenel, Ingo Fruend, Peter Yianilos,
  Moritz Mueller-Freitag, et~al.
\newblock The" something something" video database for learning and evaluating
  visual common sense.
\newblock In {\em Proceedings of the IEEE international conference on computer
  vision}, pages 5842--5850, 2017.

\bibitem{huang2020mutual}
Yifei Huang, Minjie Cai, Zhenqiang Li, Feng Lu, and Yoichi Sato.
\newblock Mutual context network for jointly estimating egocentric gaze and
  action.
\newblock {\em IEEE Transactions on Image Processing}, pages 7795--7806, 2020.

\bibitem{huang2018predicting}
Yifei Huang, Minjie Cai, Zhenqiang Li, and Yoichi Sato.
\newblock Predicting gaze in egocentric video by learning task-dependent
  attention transition.
\newblock In {\em ECCV}, pages 754--769, 2018.

\bibitem{koppula2013learning}
Hema~Swetha Koppula, Rudhir Gupta, and Ashutosh Saxena.
\newblock Learning human activities and object affordances from rgb-d videos.
\newblock {\em The International Journal of Robotics Research}, 32(8):951--970,
  2013.

\bibitem{koppula2015anticipating}
Hema~S Koppula and Ashutosh Saxena.
\newblock Anticipating human activities using object affordances for reactive
  robotic response.
\newblock {\em IEEE transactions on pattern analysis and machine intelligence},
  38(1):14--29, 2015.

\bibitem{liu2020forecasting}
Miao Liu, Siyu Tang, Yin Li, and James~M Rehg.
\newblock Forecasting human-object interaction: joint prediction of motor
  attention and actions in first person video.
\newblock In {\em ECCV}, pages 704--721. Springer, 2020.

\bibitem{luddecke2017learning}
Timo Luddecke and Florentin Worgotter.
\newblock Learning to segment affordances.
\newblock In {\em ICCVW}, pages 769--776, 2017.

\bibitem{myers2015affordance}
Austin Myers, Ching~L Teo, Cornelia Ferm{\"u}ller, and Yiannis Aloimonos.
\newblock Affordance detection of tool parts from geometric features.
\newblock In {\em ICRA}, pages 1374--1381. IEEE, 2015.

\bibitem{nagarajan2019grounded}
Tushar Nagarajan, Christoph Feichtenhofer, and Kristen Grauman.
\newblock Grounded human-object interaction hotspots from video.
\newblock In {\em ICCV}, pages 8688--8697, 2019.

\bibitem{nagarajan2020ego}
Tushar Nagarajan, Yanghao Li, Christoph Feichtenhofer, and Kristen Grauman.
\newblock Ego-topo: Environment affordances from egocentric video.
\newblock In {\em CVPR}, pages 163--172, 2020.

\bibitem{nguyen2017object}
Anh Nguyen, Dimitrios Kanoulas, Darwin~G Caldwell, and Nikos~G Tsagarakis.
\newblock Object-based affordances detection with convolutional neural networks
  and dense conditional random fields.
\newblock In {\em IROS}, pages 5908--5915. IEEE, 2017.

\bibitem{osiurak2017affordance}
Fran{\c{c}}ois Osiurak, Yves Rossetti, and Arnaud Badets.
\newblock What is an affordance? 40 years later.
\newblock {\em Neuroscience \& Biobehavioral Reviews}, 77:403--417, 2017.

\bibitem{boris_sekachev_2020_4009388}
Boris Sekachev, Nikita Manovich, Maxim Zhiltsov, Andrey Zhavoronkov, Dmitry
  Kalinin, Ben Hoff, TOsmanov, Dmitry Kruchinin, Artyom Zankevich,
  DmitriySidnev, Maksim Markelov, Johannes222, Mathis Chenuet, a andre,
  telenachos, Aleksandr Melnikov, Jijoong Kim, Liron Ilouz, Nikita Glazov,
  Priya4607, Rush Tehrani, Seungwon Jeong, Vladimir Skubriev, Sebastian
  Yonekura, vugia truong, zliang7, lizhming, and Tritin Truong.
\newblock opencv/cvat: v1.1.0, Aug. 2020.

\bibitem{selvaraju2017grad}
Ramprasaath~R Selvaraju, Michael Cogswell, Abhishek Das, Ramakrishna Vedantam,
  Devi Parikh, and Dhruv Batra.
\newblock Grad-cam:visual explanations from deep networks via gradient-based
  localization.
\newblock In {\em ICCV}, pages 618--626, 2017.

\bibitem{thermos2017deep}
Spyridon Thermos, Georgios~Th Papadopoulos, Petros Daras, and Gerasimos
  Potamianos.
\newblock Deep affordance-grounded sensorimotor object recognition.
\newblock In {\em CVPR}, pages 6167--6175, 2017.

\end{thebibliography}
}
\end{document}